\documentclass[a4paper,fleqn,9pt]{article}
\usepackage{graphicx}
\usepackage[switch]{lineno}
\usepackage{CJKutf8}

\usepackage{amsmath}
\usepackage{bm}

\usepackage{microtype}
\usepackage{hyperref}
\usepackage{graphicx}
\usepackage{subfigure}
\usepackage{booktabs} 
\newcommand{\comment}[1]{}

\title{Self-supervised denoising for massive noisy images}
\author{Feng Wang\footnote{feng.wang@empa.ch}, Trond R. Henninen, Debora Keller, Rolf Erni\\
  \small Electron Microscopy Center, EMPA, \\
  \small CH-8600, D\"{u}bendorf, Switzerland
}

\begin{document}
\maketitle


\abstract{We propose an effective deep learning model for signal reconstruction,
which requires no signal prior, no noise model calibration, and no clean samples.
This model only assumes that the noise is independent of the measurement and that the true signals share the same structured information.
We demonstrate its performance on a variety of real-world applications, from sub-\r{A}ngstr\"{o}m resolution atomic images to sub-arcsecond resolution astronomy images.}

\section{Introduction}

\begin{figure}[ht]
\vskip 0.2in
\begin{center}
\centerline{\includegraphics[width=\columnwidth]{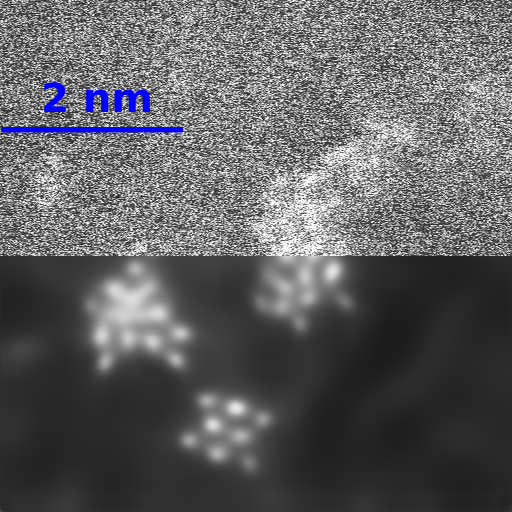}}
\caption{Example of our method restoring a high annular dark-field (HAADF) image with 512$\times$512 pixels. With dwell time less than 100 ns and a few electrons per pixel, this experimental image contains more noises than signals.  Only by using the noisy data as input, we can predict a clean atomic output.}
\label{introduction_figure}
\end{center}
\vskip -0.2in
\end{figure}

Denoising, taking the leading position in nearly every data analysis pipeline,
has recently become a popular application of deep learning\cite{tian_deep_2020}.
Traditionally, deep convolutional neural network models are either trained in a supervised fashion with paired noisy-and-clean images,
or trained in an unsupervised fashion using an unpaired dataset with the aid of a cycle training strategy.

Noise2Noise (N2N) shows that deep image denoising can be achieved only with noisy images\cite{lehtinen_noise2noise_2018}.
Although N2N still needs at least two independent noisy images for the same ground-truth image,
it has significantly alleviated the data collection pressure compared to collecting paired noisy-and-clean dataset.
This motivates the invention of recent self-supervised approaches, such as Noise2Void\cite{krull_noise2void-learning_2019}(N2V), Noise2Self\cite{batson_noise2self_2019}(N2S) and Self2Self\cite{quan_self2self_2020}(S2S), which try to learn just from noisy samples.

Our approach is directly inspirited by N2V, which uses individual noisy samples as training data, assuming that the noise is zero in mean and independent of the true signal.
To use the same noisy samples as network input and output, N2V avoids identity mapping by employing a blind-spot trick: ignoring the central pixel using a masking scheme.
Other self-supervised approaches use this very similar trade-off: sacrificing information integrity to avoid the identity mapping.
Their success comes at the price of the blind pixel(s), i.e., loss of information, either in the encoder before the first residual connection such as N2S, N2V, and in Laine's approach\cite{laine_high-quality_2019},
or in the decoding layer after the last residual connection, such as in the network designed in S2S.
However, ignoring a pixel in the noisy input or dropping out values in the middle of the network gives away some useful information, and intentionally leads to degraded performance.

We use a different approach.
The reason is that, when facing very noisy datasets, as is shown in Fig.\ref{introduction_figure}, we cannot afford any information loss.
We stick to the information integrity, at a price that we have to drop all the residual connections to avoid identity mapping.

The conventional architectures, such as U-Nets\cite{ronneberger_u-net_2015} and residual networks\cite{mao_image_2016,zarshenas_deep_2018,shi_hierarchical_2019, zhang_residual_2020}, have been selected to build up many denoising models.
These residual learning models use residual connections to transport low-frequency features and fine details.
When trained on noisy dataset only, they give out useless identical outputs.
Instead, when trained on a noisy--clean paired dataset, they can produce excellent denoising results.
As we are matching the same input and output, we cannot use residual connections in our network.
Thus, we turn our focus to the autoencoder, as it is GPU memory friendly designed.
This architecture has provided a general framework for lossy compression followed by decompression of the data to solve the problem of identical mapping.
To accelerate the convergence of the autoencoder, we redesign the architecture in the decoder by adopting the multi-resolution convolutional neural network\cite{wang_multi-resolution_2020} (MCNN): matching additional outputs at all frequencies in pursuit of better stability and faster convergence.
A typical convergence curve for our model is demonstrated in Fig.\ref{wqy_loss_figure}.
Besides, we also test the transfer learning technique, by adopting VGG19\cite{simonyan_very_2015} as the encoder and only training the parameters in the decoders. But unfortunately we do not find better convergence than MCNN.

Finally, we note that Vincent et al.\cite{vincent_stacked_2010} uses supervised stacked denoising autoencoders to map samples to a low-dimensional representation then back to the original space,
but their purpose is to learn robust representations for classification by adding noise.

\section{Method}

The field of image denoising is currently dominated by supervised deep learning models trained on pairs of noisy--and--clean image sets.
These models are mostly developed for the enhancement of the image quality of commercial cameras, and usually start with samples recorded at a high signal-to-noise ratios (SNR).
Modern fast detectors can easily produce hundreds of terabytes of images per hour\cite{spurgeon_towards_2020}, but unlike common commercially available cameras, this kind of detectors often give out results containing more noise than signals.
Like many practical applications, it is not possible to obtain ground-truth images to train a supervised denoising algorithm.
Also, due to the complex experimental configurations and mathematical models behind the experiment, simulating ground-truth images is very difficult, and training effective unsupervised models are very challenging or even impossible.
However, despite the heavy data stream intensity, we can assume that the content of the images in the same experiment falls into the same domain.
By assuming that the acquired dataset contains the same structured information, we can randomly sample a subset from the recorded samples, and train a self-supervised mode.

A simple principle in the denoising theory is: if we can replace a pixel's value with, for example, the average value of $n$ nearby pixels,
the variance law ensures that the standard deviation of the average is divided by a factor of $\sqrt{n}$.
Therefore, if we can find for each pixel $N$ other pixels in the same image with the same pixel value,
we can divide the noise by $\sqrt{N}$.
As a consequence of this principle, different similar pixels/blocks searching and weighting strategies give rise to a series of denoising methods achieved by averaging,
such as Non-local means\cite{buades_non-local_2005} (NLM) and Block-matching and 3D filtering\cite{dabov_image_2007} (BM3D).

As only the noisy input images are available, these block/pixel-based methods that search for a small number of similar patches in their neighborhoods, usually only work in the current image rather than in the whole dataset and are very slow. On the other hand, deep learning-based approaches can implicitly match arbitrarily large patterns from all the samples.






The deep networks that are trained by the standard gradient descent algorithm are memorizing the training data, and use the memorized samples for prediction with a weighted similarity function\cite{webster_this_2021,domingos_every_2020,zhang_understanding_2021,noauthor_sgd_nodate,song_machine_2017,carlini_secret_2019}.

For our self-supervised model $x = f_\theta(x)$, in which $\theta$ are trainable parameters and $f_\theta(x)$ is not a simple identity function,
if we write the loss at sample $x_i$ as $l(x_i, f_\theta(x_i))$, then the accumulated loss can be defined as
\begin{equation}
\mathcal{L}(\theta) = \sum_i l(x_i, f_\theta(x_i)). \label{eq_loss}
\end{equation}

Training this model using a gradient descent optimizer, the update of $\theta$ at step $n+1$ is
\begin{equation}
\theta_{n+1} = \theta_n - \gamma \nabla_\theta \mathcal{L}(\theta_n). \label{eq_gd}
\end{equation}

When the step size $\gamma$ is small enough, Eq.\ref{eq_gd} is a forward Euler procedure for solving an ordinary differential equation (ODE)
\begin{equation}
\frac{\mathop{d\theta}}{\mathop{dt}} = \frac{\theta_{n+1}-\theta_n}{\gamma} = - \nabla_\theta \mathcal{L}(\theta) = - \frac{\partial{\mathcal{L}(\theta)}}{\partial{\theta}}. \label{eq_ode}
\end{equation}

Substituting Eq.\ref{eq_loss} in Eq.\ref{eq_ode}, we have
\begin{align}
\frac{\mathop{d\theta}}{\mathop{dt}} & = - \frac{\partial{\mathcal{L}(\theta)}}{\partial{\theta}} \\
                           & = - \sum_i \frac{ \partial{ l(x_i, f_\theta(x_i)) } }{\partial{\theta}} \\
                           & = - \sum_i \frac{ \partial{ l(x_i, f_\theta(x_i)) } }{\partial{f_\theta(x_i)}} \frac{ \partial{f_\theta(x_i)} }{\partial{\theta}}.
\end{align}

Considering the changes of our self-supervised model $f_\theta(x)$ over the training procedure, we can derive
\begin{align}
\frac{\mathop{df_\theta(x)}}{\mathop{dt}}
& = \sum_j \frac{\partial{f_\theta(x)}}{\partial{\theta_j}} \frac{\mathop{d\theta_j}}{\mathop{dt}} \\
& = - \sum_j \frac{\partial{f_\theta(x)}}{\partial{\theta_j}} \sum_i \frac{ \partial{ l(x_i, f_\theta(x_i)) } }{\partial{f_\theta(x_i)}} \frac{ \partial{f_\theta(x_i)} }{\partial{\theta_j}} \\
& = - \sum_i \frac{ \partial{l(x_i, f_\theta(x_i)) } }{ \partial{f_\theta(x_i)} } \sum_j \frac{ \partial{f_\theta(x)} }{ \partial{\theta_j}  } \frac{ \partial{f_\theta(x_i)} }{ \partial{\theta_j} } \\
& = \sum_i \alpha_\theta(x_i) \, \kappa_\theta(x, x_i) ,
\end{align}
in which $\alpha_\theta(x_i)= - \frac{ \partial{l(x_i, f_\theta(x_i)) } }{ \partial{f_\theta(x_i)} } $ and $\kappa_\theta(x, x_i) = \sum_j \frac{ \partial{f_\theta(x)} }{ \partial{\theta_j}  } \frac{ \partial{f_\theta(x_i)} }{ \partial{\theta_j} }$.

After finishing the training at time step $T$, this self-supervised model characterized by the trainable parameters $\theta_T$ becomes
\begin{align}
f_{\theta_T}(x)
& = f_{\theta_0}(x) + \sum_i \int_0^T \alpha_{\theta_{t}}(x_i) \, \kappa_{\theta_t}(x, x_i)  \mathop{dt} \\
& = f_{\theta_0}(x) + \sum_i \frac{ \int_0^T \alpha_{\theta_t}(x_i) \, \kappa_{\theta_t}(x, x_i) \mathop{dt} }{ \int_0^T \kappa_{\theta_t}(x, x_i) \mathop{dt} } \int_0^T \kappa_{\theta_t}(x, x_i)  \mathop{dt} \\
& = b_0 + \sum_i \bm{\alpha}_i(x) \bm{K}(x, x_i), \label{eq_query}
\end{align}

in which $b_0 = f_{\theta_0}(x)$ is determined by $\theta_0$ initialized by algorithms such as Glorot uniform\cite{glorot_understanding_2010}, and usually falls into a normal distribution as the behaviour of this network is mathematically equivalent to a Gaussian process\cite{lee_deep_2018}. And
$\bm{\alpha}_i(x) = \frac{ \int_0^T \alpha_{\theta_t}(x_i) \, \kappa_{\theta_t}(x, x_i) \mathop{dt} }{ \int_0^T \kappa_{\theta_t}(x, x_i) \mathop{dt} }$ is a weighted term of $\kappa_{\theta_t}(x, x_i)$ over training time $t$,
and $\bm{K}(x, x_i)$ is a kernel function measuring similarities between the input sample $x$ and memorized samples $x_i$.

\begin{figure}[ht]
\vskip 0.2in
\begin{center}
\centerline{\includegraphics[width=\columnwidth]{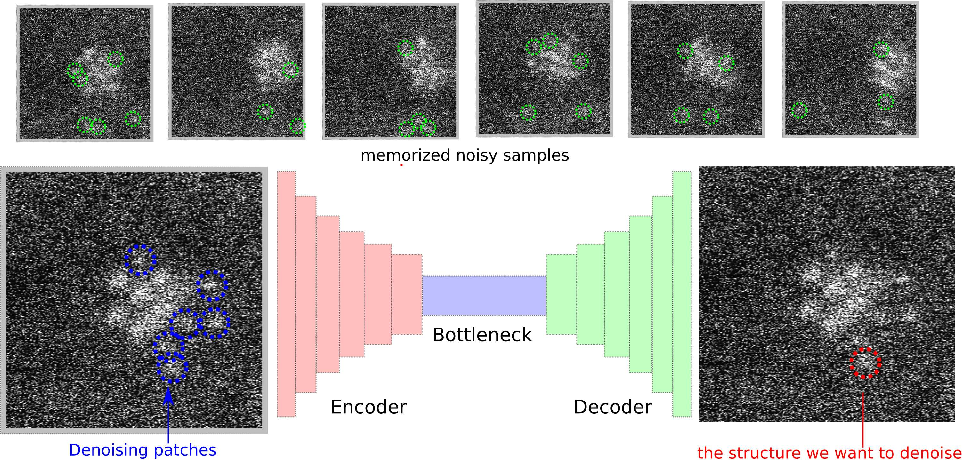}}
\caption{The idea behind our model: querying learned samples to generate a weighted output.}
\label{idea_figure}
\end{center}
\vskip -0.2in
\end{figure}

Eq.\ref{eq_query} indicates that when making prediction for an input sample $x$, our model queries all the learned samples in the memory, and generates a weighted averaged output, as is shown in Fig.\ref{idea_figure}.
There is no additional constraint on the model selection from this equation. For the sake of memory efficiency, we use a compressive auto-encoder as our backbone network.

\section{Experiments}

We evaluate our model with several denoising tasks: images of simulated Chinese-Japanese-Korean (CJK) character at a SNR around -10 dB, fluorescence microscopy dataset of Tribolium embryo,
scanning transmission electron microscopy (STEM) images of atoms and nanoparticles in ionic liquids, and astronomy images.

\subsection{Channels in the bottleneck layer}

Most of the denoising methods have hyper--parameters controlling the threshold of the denoising, for example, the rank of the matrices, the size of the filters, and even the prior knowledge of the noise variance.
We use the channels of the bottleneck layer to control the degree of denoising.
With a large number of channels we assume the experimental dataset has a lot of detailed signals to restore;
with a number of channels as small as 1, i.e. compressing an input of dimension $128 \times 128 \time 1$ to $8 \time 8 \time 1$, we assume a very sparse dataset and that there are only a few structure patterns in the signal.

\begin{figure}[ht]
\vskip 0.2in
\begin{center}
\centerline{\includegraphics[width=\columnwidth]{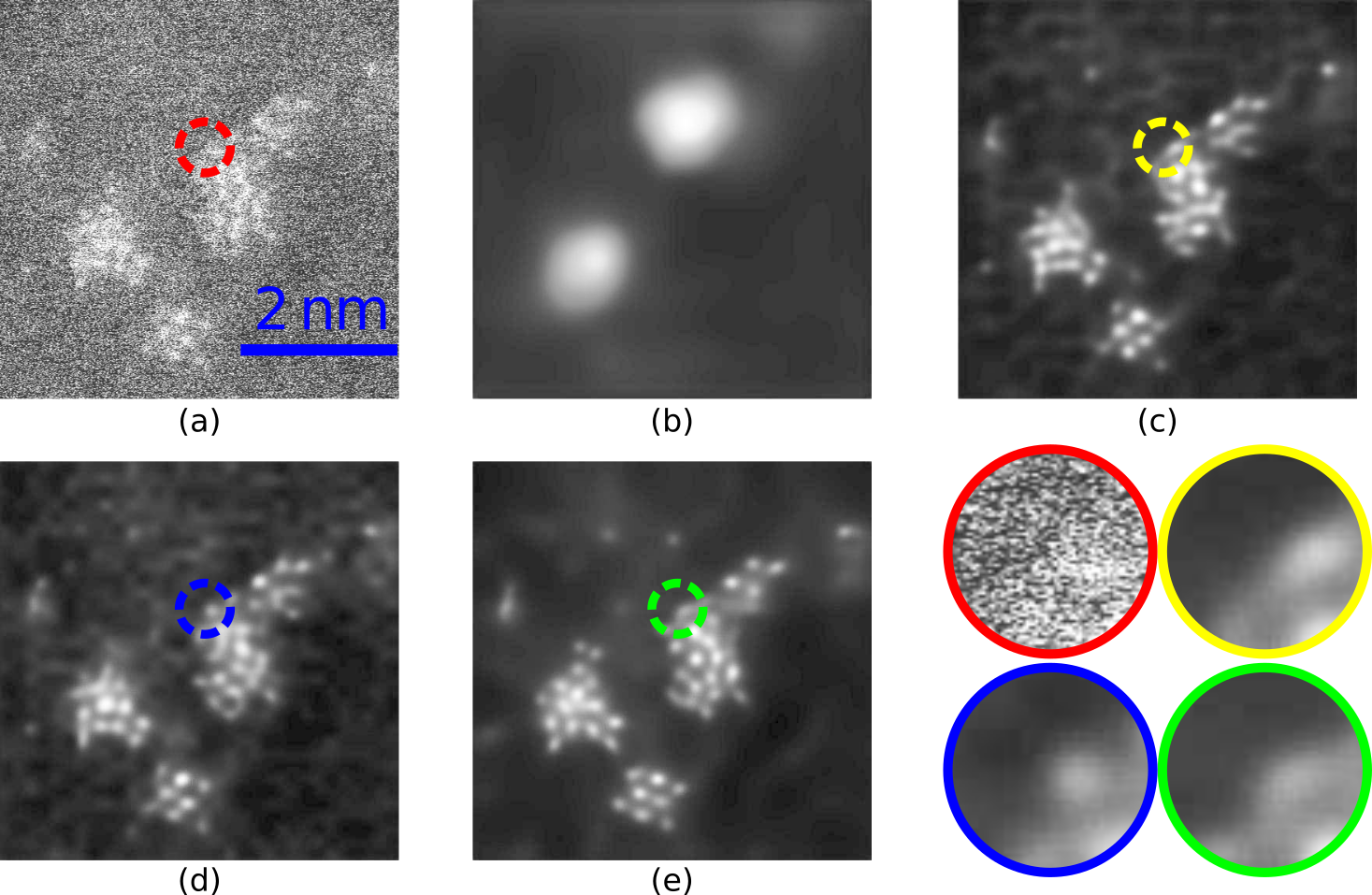}}
\caption{Examining the convergence speed and effect of the bottleneck channels on an experimental image with 512$\times$512 pixels shown in (a). At epoch 30, with a bottleneck layer of dimension 16$\times$16$\times$1 in the model, only the outline of two big clusters are predicted, as is shown in (b). However, at epoch 130, almost all the atoms are restored, as is shown in (e). While with a wider bottleneck in the model, for example, 16$\times$16$\times$4 in (c) and 16$\times$16$\times$16 in (d), the atoms are quickly restored at epoch 30. Also, the model with fewer bottleneck channels in (e) gives a much smoother background than the other ones in (c) and (d).}
\label{bottleneck_figure}
\end{center}
\vskip -0.2in
\end{figure}

We examine how our model behaves as we increase the channels in the bottleneck layer.
Fig.\ref{bottleneck_figure} (b), (c) and (d) show the denoising results of the same experimental image (a) at epoch 30 while increasing the bottleneck channels from 1 to 16, using 1000 noisy images with 512$\times$512 pixels:
the more bottleneck channels the model has, the quicker the network converges, and the finer details are passed to the output.
However, the model with one channel still makes a good prediction after a long run,
as is shown in Fig.\ref{bottleneck_figure}(3). After 130 epochs in two hours, this model reveals nearly all the atoms.
This means that we can define the degree of fine details by setting up different bottleneck channels.
However, to save the training time, all the models presented in the context stick to 64 bottleneck channels, unless otherwise specified.

\subsection{Reliability of image restoration}

\begin{figure}[ht]
\vskip 0.2in
\begin{center}
\centerline{\includegraphics[width=\columnwidth]{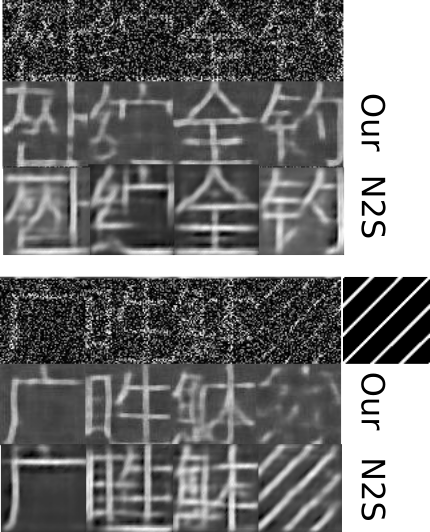}}
\caption{Our model and N2S working on heavily noisy Chinese-Japanese-Korean (CJK) character images (SNR around -10 dB), and give out similar denoising results.}
\label{cjk_figure}
\end{center}
\vskip -0.2in
\end{figure}

\begin{figure}[ht]
\vskip 0.2in
\begin{center}
\centerline{\includegraphics[width=\columnwidth]{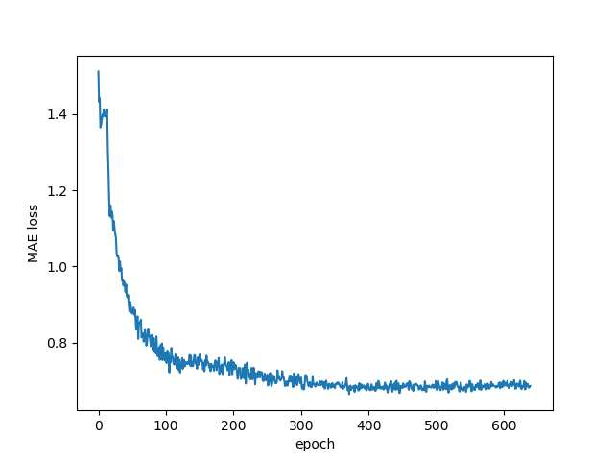}}
\caption{On mixed CJK characters, our model quickly converges quickly on the testing set. This fast convergence speed comes from MCNN.}
\label{wqy_loss_figure}
\end{center}
\vskip -0.2in
\end{figure}

To demonstrate the performance of our model on the heavily noisy dataset,
we extracted 38,631 unique Chinese-Japanese-Korean (CJK) images from open source font Wen Quan Yi\cite{wqy} (WQY), rescaled these images to 64$\times$64 pixels, normalized their intensities to the range $[0, 1]$,
added uniform noise in the range $[-2, 2]$ to each pixel, reset negative intensities to 0, and then renormalized their intensities to the range $[0, 1]$.
We finally generated a noisy dataset with a SNR close to -10 dB, and tested our model and N2S on this dataset.
We spent about two hours training our model and N2S on this dataset for 200 epochs.
Our model has four layers in bottleneck, i.e., we compress a noisy image which has $64 \times 64 \times 1$ pixels to $4 \times 4 \times 4$ pixels.
We demonstrate the restored results for eight characters in Fig.\ref{cjk_figure}.
The first seven characters are correctly reconstructed by our method and N2S, even though their corresponding noisy inputs are difficult to read even for a native reader.
The eighth one is from a dummy character.
In WQY, this character is composed of five evenly-spaced thin anti-diagonal strokes.
These anti-diagonal strokes are illegal for CJK characters.
Our model fails to make a correct restoration on this case, even if we expand the bottleneck threshold to 64, as it cannot learn such a pattern from the input images, but N2S interestingly can pick it up.
On other characters, our model shows similar restoration performance as N2S.
The convergence curve for L1 loss is presented in Fig.\ref{wqy_loss_figure}.


\begin{figure}[ht]
\vskip 0.2in
\begin{center}
\centerline{\includegraphics[width=\columnwidth]{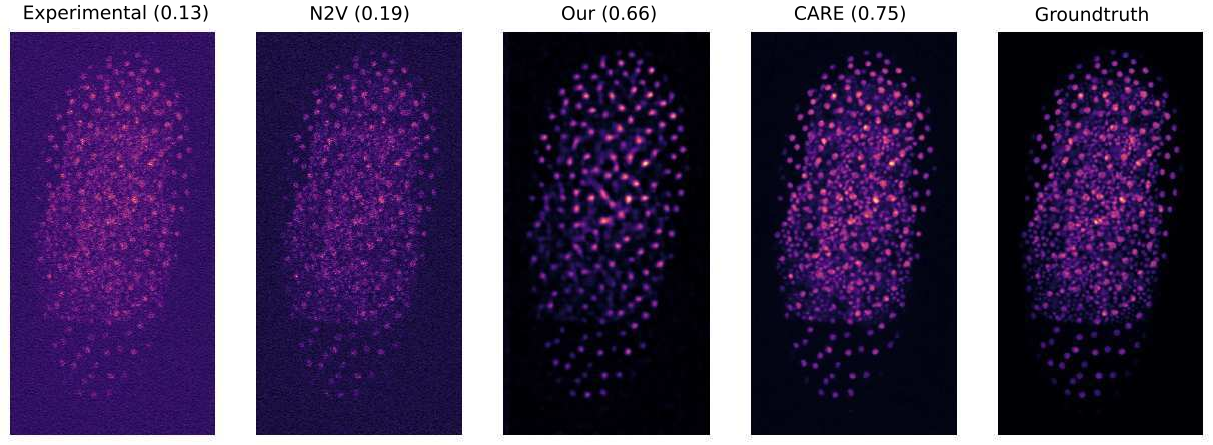}}
\caption{Stacked Tribolium dataset (45 images in total). From left to right: maximum projection of low-SNR images (SSIM=0.13), maximum projection of Noise2Void (self-supervised) denoised images (SSIM=0.19), maximum projection of our model (self-supervised) denoised images (SSIM=0.66), maximum projection of CARE (supervised) denoised images (SSIM=0.75) and maximum high-SNR images (GT). The structure similarities (SSIM) are presented on the top of each figure.}
\label{compare_n2v_n2p_se}
\end{center}
\vskip -0.2in
\end{figure}

Even though our model is intentionally designed for denoising heavily noisy images, we still can compare its performance with the supervised content-aware image restoration (CARE) network\cite{weigert_content-aware_2018}.
CARE comes with a low-SNR and high-SNR 3D dataset of Tribolium.
The low-SNR dataset is composed of 45 images with 954$\times$486 pixels, in which the recorded intensities are very weak and the signals are very sparse.
The stacked 45 low-SNR images are presented in Fig.\ref{compare_n2v_n2p_se}.
As the dataset is very small, we expanded the dataset by rotation and flipping, and spent a few hours training the model for 136 epochs.
We find our model produces a good interpretation: its SSIM (0.66) is already very close to the SSIM produced by CARES (0.75), as is demonstrated on the top of each figure in Fig.\ref{compare_n2v_n2p_se}.
For comparison, we train Noise2Void (N2V) for 200 epochs using their suggested architecture on the same dataset, and find that its performance is not as good as ours.

\subsection{Sub-\r{A}ngstr\"{o}m resolution STEM images}

High spatial and temporal resolution visualization is essential in accessing dynamic particle formation processes in liquid at the atomic scale\cite{keller_atomic_2020,henninen_structure_2020}.
However, minimizing the electron dose while recording with a simultaneously high spatial and temporal resolution leads to very noisy images, let along the high background signals coming from tens of nanometers of carbon support and ionic liquid.

Molecular dynamics could be employed to simulate the physical process but it is very expensive or nearly impossible for a large number of atoms , and it would have to rely on physical models.
It is therefore difficult to simulate \textit{ground-truth} images for training a supervised model.
Unsupervised and some self-supervised approaches are also difficult\cite{wang_noise2atom_2020} in interpreting atomic clusters with a wide number of atoms ranging from one to several thousands.

\begin{figure*}[ht]
\vskip 0.2in
\begin{center}
\centerline{\includegraphics[width=1.0\columnwidth]{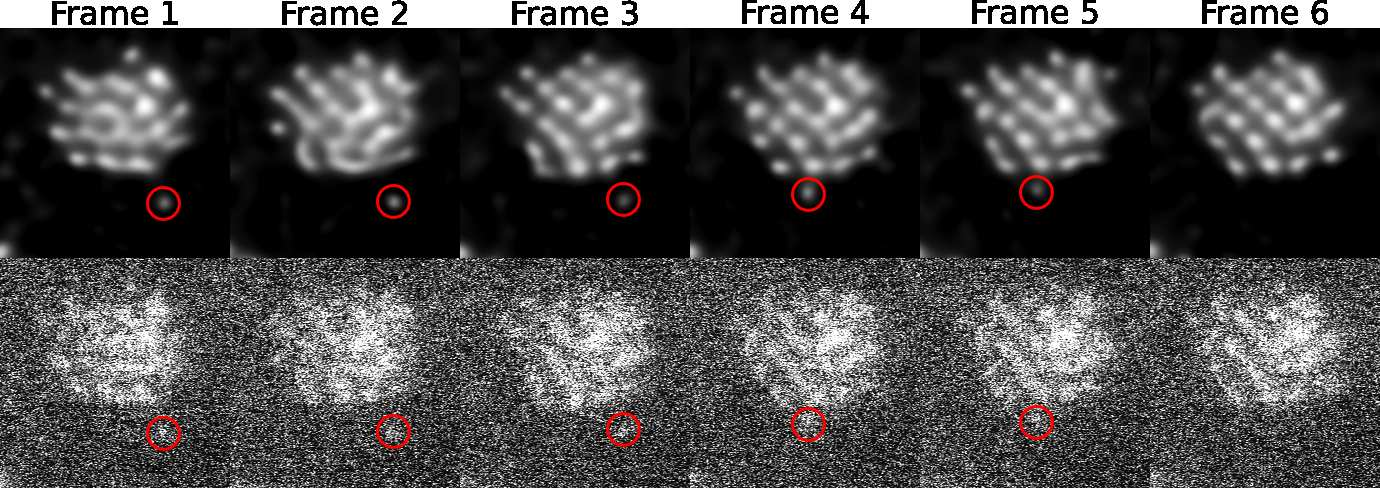}}
\caption{Our model allows atomic dynamics: living view of an isolated atom (in red circle) gets absorbed by a nearby cluster. The time resolution is 15 fps.}
\label{compare_s9_1}
\end{center}
\vskip -0.2in
\end{figure*}

\begin{figure*}[ht]
\vskip 0.2in
\begin{center}
\centerline{\includegraphics[width=1.0\columnwidth]{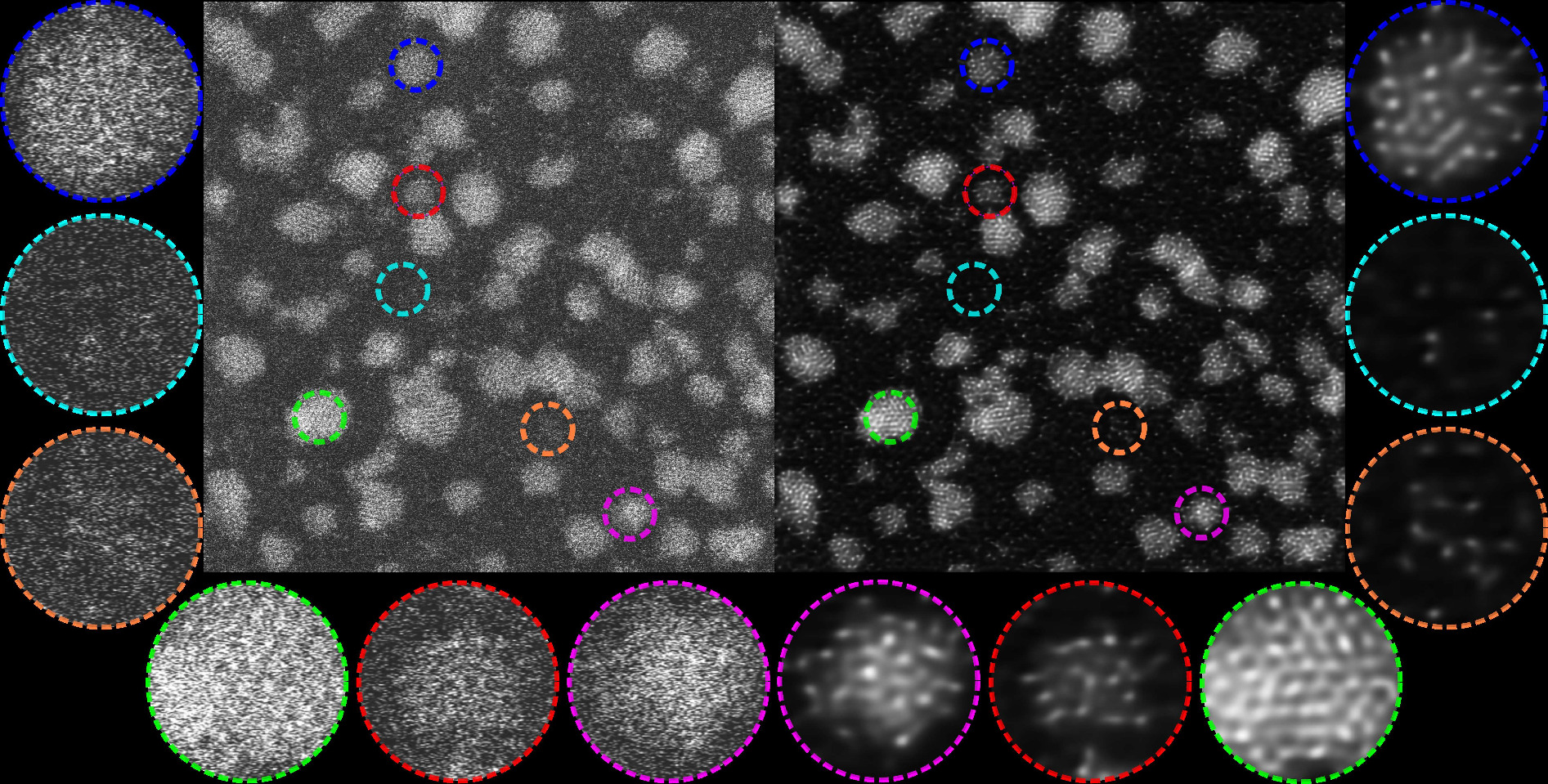}}
\caption{Our model helps the interpretation of isolated atoms and huge atomic clusters containing hundreds of atoms simultaneously. Left: experimental image; right: denoised result.}
\label{compare_debora}
\end{center}
\vskip -0.2in
\end{figure*}

We test our model on various experimental STEM images recorded using an electron dose ranging from $10^{4} e $\r{A}$^{-2}$ to $10^{6} e $\r{A}$^{-2}$.
The images are acquired with typical pixel sizes of 6.3-12.6 pm at 1-20 frames per second (fps) and a frame size from $512 \times 512$ pixels to $2048 \times 2048$ pixels.
Each dataset includes thousands of images, and it takes about 5 hours to train our model for 100 epochs with a dataset of $1000 \times 512 \times 512$ pixels.

our model produces clear and consistent atomic images on a dataset acquired at 15 fps:
as is shown in Fig.\ref{compare_s9_1}, we can see the process of an isolated atom is being absorbed by a rotating and reconstructing cluster.
For the second example, as is shown in Fig.\ref{compare_debora}, our model produces a very good interpretation of the isolated atoms, which is barely visible by human eyes,
and on the overlapped atomic clusters, which can be classified by their orientations.
In another words, our model allows for studying individual atoms in real-time based on noisy experimental data.

\begin{figure}[ht]
\vskip 0.2in
\begin{center}
\centerline{\includegraphics[width=\columnwidth]{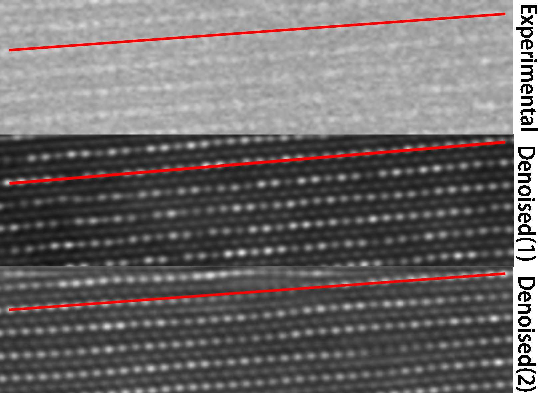}}
\caption{For the long range features, the model training with larger image patches produces better results than the one trained with smaller patches. The model Denoised(1) is trained using image patches with $1024 \times 1024$ pixels, and this model correctly predicts straight lattices. The model Denoised(2) is trained using image patches with $128 \times 128$ pixels, and in some of the weak intensity area, it predicts lattices with some distortion.}
\label{_lattice_1_figure}
\end{center}
\vskip -0.2in
\end{figure}

\begin{figure}[ht]
\vskip 0.2in
\begin{center}
\centerline{\includegraphics[width=\columnwidth]{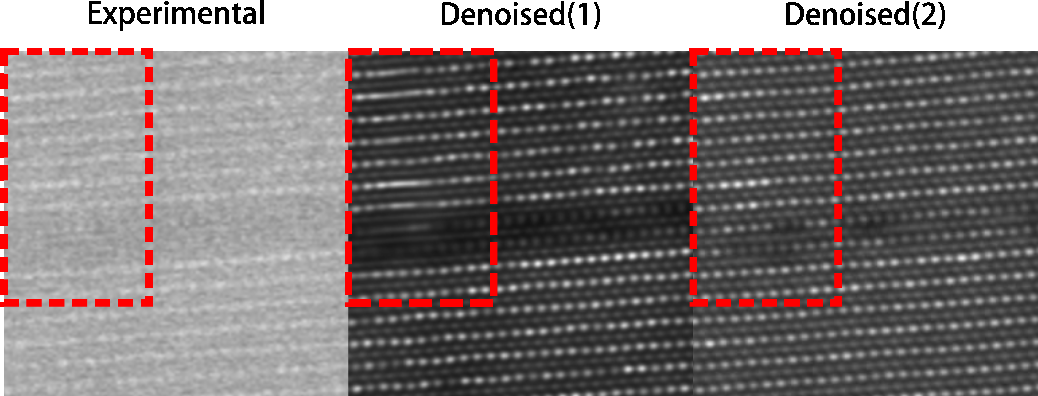}}
\caption{For the local area features, the model training with smaller image patches produces better results than the one trained with larger patches. The model Denoised(1) is trained using image patches with $1024 \times 1024$ pixels, and this model gives blurry results on some of the neighbouring atoms. The model Denoised(2) is trained using image patches with $128 \times 128$ pixels, it predicts good results with every atom explicitly reconstructed even in some of the weak intensity area.}
\label{bend_2_figure}
\end{center}
\vskip -0.2in
\end{figure}

And interestingly, our model shows how the reconstruction results are affected by the samples it memorized: training with small image patches, this model focuses on local range features, while training with large image patches, this model handles long range features better, as is demonstrated in SmCo Magnetic Block Permanent Magnet images in Fig.\ref{_lattice_1_figure} and Fig.\ref{bend_2_figure}.

\subsection{Sub-arcsecond resolution astronomy images}

\begin{figure*}[ht]
\vskip 0.2in
\begin{center}
\centerline{\includegraphics[width=1.0\columnwidth]{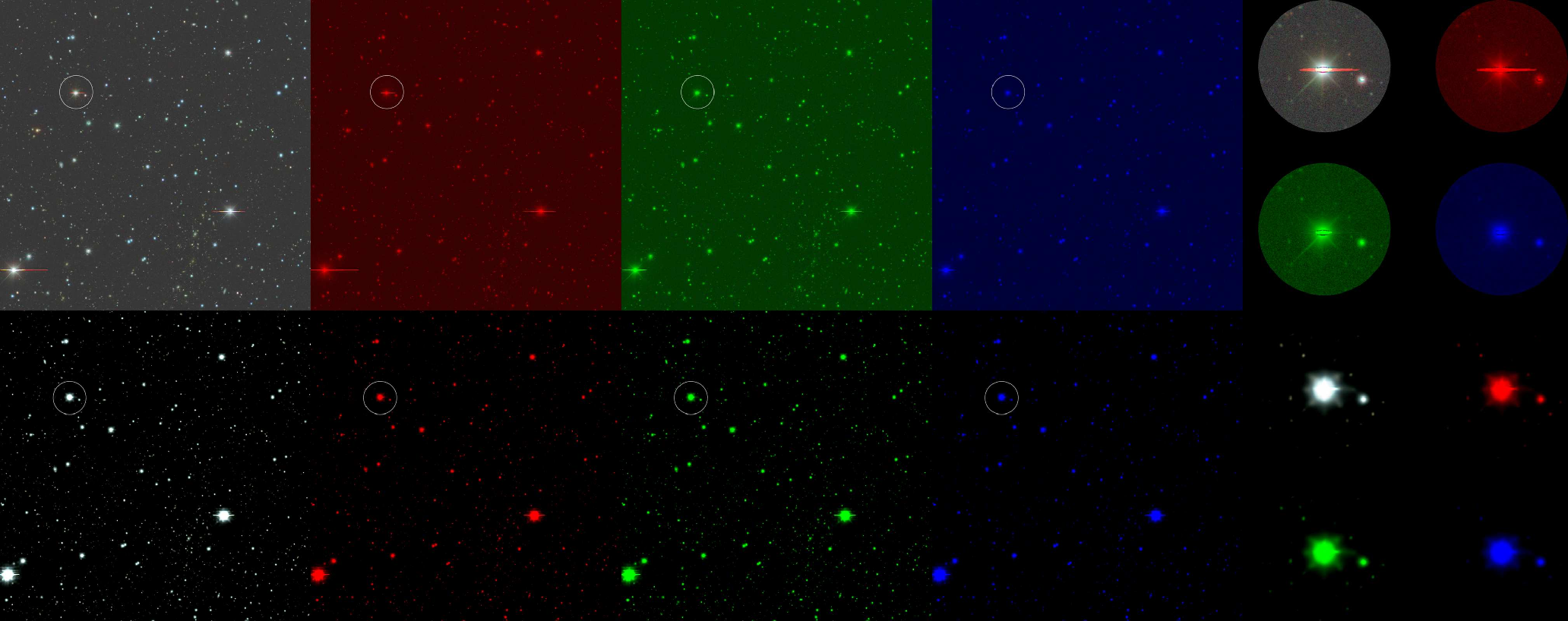}}
\caption{Our model corrects most of the noises for DR8 dataset. Upper row: the raw image; lower row: the restored image. From left to right: full RGB channel, red channel, green channel, blue channel and the amplification of selected zones. Zoom in to see more details.}
\label{dr9_figure}
\end{center}
\vskip -0.2in
\end{figure*}

The DESI legacy imaging surveys\cite{dey_overview_2019} tries to image about 14,000 deg$^2$ of the extragalactic sky visible from the northern hemisphere using telescopes.
A combination of three telescopes is used to provide optical imaging for the legacy surveys.
Denoising takes the leading position in the data processing pipeline, to restore common defects such as bad or over-saturated pixels, ghost patterns, sky-gradients, and striping noise.
We collect 16 bricks starting from DR8-south 1410m032 to 1416p230. Each brick covers 0.25 $\times$ 0.25 deg$^2$ using $3600 \times 3600 \times 3$ pixels, with a pixel resolution of 0.262 arcsecond.
We randomly sample 12288 patches from the dataset, each patch contains $256 \time 256 \times 3$ pixels, and spend about 16 hours to train our model for 136 epochs with a batch size of 16.
Our model produces clear results while maintaining the intensities in all three channels and removing most of the image defects.
As is shown in Fig.\ref{dr9_figure}, our model removes nearly all the background noise, fixes over-saturated pixels and largely corrects stripping noise.

\section{Conclusion}

We have demonstrated a general framework for image denoising, which only uses the noisy images as input and has no prerequisite on the training set nor the prior information.
Our approach is dedicated to restoring very noisy images for ballooning datasets produced by modern detectors.
Instead of applying the popular dropout/masking pixel(s) tricks,
we adopt a compressive autoencoder architecture to reserve all the information from the training data,
and adopting MCNN technique for fast convergence.
This is a very different approach compared to other self-supervised models: these other models give away some useful information either before or during the training process, risking a reduced performance.
Experiments show that our approach can make a good interpretation from a variety of noisy samples, ranging from sub-\r{A}ngstr\"{o}m resolution atomic images to sub-arcsecond resolution astronomy images.
We hope our framework will find a lot of applications in a wide range of fields, especially where there is abundantly available measurement data, which is very noisy as the detector signal being strictly limited, e.g. due to the short exposure time or the weak signal source.

\section*{Acknowledgement}

We gratefully acknowledge the support of Xiaoke Mu from Institute of nanotechnology (INT) and Karlsruhe Nano Micro Facility (KNMF) in Karlsruhe Institute of Technology (KIT) for the discussion and sharing the experimental images.

\bibliography{main}
\bibliographystyle{unsrt}

\end{document}